\documentclass[letterpaper]{article}

\usepackage{natbib,alifeconf}
\usepackage{amsmath}
\usepackage{amssymb}
\usepackage{url,hyperref,cleveref}
\usepackage{booktabs}
\usepackage{mdframed}

\usepackage{multirow}
\usepackage{float}
\usepackage{tikz}
\graphicspath{{figures/}}

\newcommand\blfootnote[1]{%
  \begingroup
  \renewcommand\thefootnote{}\footnote{#1}%
  \addtocounter{footnote}{-1}%
  \endgroup
}

\title{From Signals to Structure: How Memory Architecture\\
Drives Language Emergence in LLM Agents}

\author{
    Yashar Talebirad$^{1}$,
    Eden Redman$^{2}$,
    Ali Parsaee$^{1}$, \and
    Osmar R. Za\"{i}ane$^{1}$ \\
    \mbox{}\\
    $^1$Alberta Machine Intelligence Institute, University of Alberta, Edmonton, Canada \\
    $^2$Network for Applied Technology, Edmonton, Canada \\
    talebira@ualberta.ca, eden@nat.ltd
}

\begin{document}
\raggedbottom
\maketitle

\begin{abstract}
How do two agents invent a shared language from scratch?
In a Lewis signaling game, a sender and receiver must coordinate on a code using only their interaction history.
We study five memory architectures across varying channel configurations with LLM agents and find that memory architecture matters more than channel capacity.
Agents with a persistent private notebook benefit from surplus channel capacity and avoid the high-capacity collapse seen in stateless agents, achieving the most reliable coordination ($0.867 \pm 0.023$ at capacity $= 25$).
Stateless agents peak at moderate capacity and then degrade as the vocabulary grows beyond what a rolling context window can track.
The notebook externalizes learned conventions, freeing agents from having to re-derive codes each round.
An information bottleneck-inspired argument predicts an optimal capacity equal to the number of objects.
Instead, the bottleneck (capacity $= 8$) proves to be a fragility point, and surplus capacity is generally better.
We show that channel capacity alone cannot predict coordination; memory architecture determines whether agents turn interaction history into stable conventions, and both dimensions are needed to understand how signals become language.
\end{abstract}

\vspace{-14pt}

\blfootnote{\textcopyright 2026 Yashar Talebirad, Eden Redman, Ali Parsaee, and Osmar R. Za\"{i}ane. Published under a Creative Commons Attribution 4.0 International (CC BY 4.0) license.}
\blfootnote{This is the authors' version of a paper accepted to ALIFE 2026, with minor aesthetic changes from the version of record, which appears in the ALIFE 2026 conference proceedings.}

\section{Introduction}

The Lewis signaling game \citep{lewis1969convention} is a minimal model of communication emergence: a sender observes a designated target among a set of candidates and transmits a constrained signal, and a receiver sees the same candidates and the signal, then identifies the target.
No pre-negotiated meanings exist, and agents converge through repeated coordination alone.
Consistent success in this coordination task means that a new language has been invented by the agents.
Because the objects are described by familiar features and the agents are pretrained models with semantic priors, this language is a mapping from an arbitrary signal space onto an already structured meaning space.

Large language models introduce a different kind of agent.
Unlike gradient-trained agents, LLMs provide a general-purpose reasoner that can be placed in a wide range of simulated environments without retraining, bringing linguistic and inferential priors to each task.
LLMs also have the ability to adapt through \emph{in-context learning} \citep{brown2020language}: they can reason over a history of prior interactions to refine strategy on each new call.
This shifts attention from model architecture to \emph{memory architecture}: what information each agent retains across rounds and in what form.
The scratchpad \citep{nye2021scratchpads} and chain-of-thought \citep{wei2022chain} literatures show that the structure of intermediate representations shapes what LLMs can compute.
In a signaling game, how an agent stores what it has learned determines the language it can invent.
Classical emergent communication research uses this framework with gradient-trained neural agents, finding that compositional protocols, those in which signal structure mirrors object structure, emerge under the right pressures \citep{lazaridou2017multi}.
Information-theoretic bottleneck arguments \citep{tishby1999information} motivate paying special attention when the channel is scarce relative to the number of referents.
\citet{resnick2020capacity} study compositionality as a function of both bandwidth and model capacity.
The channel is most stressed when its capacity (cap), the number of distinct messages available, exactly matches the number of objects (here, cap $= 8$).
Below that floor, agents cannot distinguish all objects; above it, compression pressure to reuse structure weakens.
Whether this cap~$= 8$ point behaves as a compositional optimum for LLM agents, or whether coordination turns less on the channel than on what agents can remember, is what this paper sets out to answer.

We run three studies, all with \texttt{gpt-5.4-mini} as the base model for both agents.
Study~1 compares the five memory architectures at a fixed channel, Study~2 sweeps channel capacity from 4 to 125 by varying vocabulary size $|V|$ and message length $L$, and Study~3 separates consolidation from history length.
We show that channel capacity alone cannot predict whether agents will coordinate at the bottleneck.
The information bottleneck at cap $= 8$ turns out to be a fragility point rather than a compositional optimum.

The capacity-only framing treats performance as a property of the channel and omits memory architecture: whether agents can write down what they have learned and carry it forward rather than re-deriving it each round.
Agents with a persistent private notebook benefit from surplus capacity without the high-capacity collapse, while stateless agents peak at moderate capacity and then degrade as the code space grows beyond what a rolling window can track.
We show that memory architecture reshapes the capacity-performance curve rather than merely shifting it.

\section{Background}

\paragraph{Lewis signaling games.}
Lewis games have been studied analytically \citep{skyrms2010signals}, computationally \citep{kirby2001spontaneous}, and in human experiments \citep{kirby2008cumulative}.
The iterated learning tradition shows that transmission pressure can drive the emergence of compositional structure.
More specifically, \citet{kirby2015compression} argue that compositionality requires both communication pressure (discriminability) and compression pressure (learnability); neither alone suffices.
This dual-pressure account provides the theoretical ground for our comparison of memory architectures.

\paragraph{Emergent communication with neural agents.}
\citet{lazaridou2017multi} established the modern deep learning framework for referential games; sender-receiver pairs trained with the REINFORCE algorithm develop protocols that are functional but often non-compositional \citep{lowe2019pitfalls}.
Compositionality, measured via topographic similarity \citep[TopSim;][]{brighton2006understanding}, emerges more reliably under structured input spaces \citep{lazaridou2018referential}, iterated learning pressure \citep{ren2020compositional}, or ease-of-teaching objectives \citep{li2019ease}.
Furthermore, \citet{resnick2020capacity} identify channel capacity as a key variable, arguing for an optimal bandwidth range.
All of this prior work uses gradient-trained agents.
In contrast, we study frozen LLM agents whose only adaptation mechanism is in-context reasoning.

\paragraph{LLMs as communicating agents.}
The most directly related prior work is \citet{kouwenhoven2025searching}, who run LLMs in an iterated referential game with generational transmission, finding that initially holistic languages, where each signal names a whole object, acquire compositional structure across generations.
In our design, rather than passing language between generations, agents accumulate memory within a single run.
\citet{ashery2025emergent} show that LLM populations spontaneously develop shared naming conventions, confirming that convention formation dynamics are not unique to humans or gradient-trained systems.
On the coordination side, \citet{akata2025playing} find that LLMs perform poorly in pure coordination games unless some mechanism breaks the symmetry between agents.
\citet{talebirad2025loopbench} report a similar pattern in a distributed graph-coloring benchmark: agents can loop indefinitely without a way to pass strategies, and they escape deadlock only when memory structures support emergent symmetry breaking.
These results suggest that memory architecture may be the key factor separating LLM agents that coordinate from those that do not.
None of these studies, however, treats memory architecture as a controlled variable or pairs it with channel capacity, as we do here.

\section{Experimental Setup}

\begin{figure}[t]
\centering
\begin{tikzpicture}[
    arr/.style  = {->, >=stealth, semithick},
    darr/.style = {->, >=stealth, dashed, gray, semithick}
]
\draw[rounded corners=5pt, fill=gray!7, draw=black!50]
    (-1.15,-0.5) rectangle (1.15,2.7);
\node[font=\small\bfseries, align=center] at (0,3.1)
    {Agent A \\ {\normalfont\scriptsize(sender)}};
\draw[fill=orange!25, draw=black!50, rounded corners=2pt]
    (-0.5,2.11) rectangle (0.5,2.49);
\fill[red!65] (0,2.30) circle (0.10cm);
\draw[fill=white, draw=black!35, rounded corners=2pt]
    (-0.5,1.56) rectangle (0.5,1.94);
\fill[blue!60] (-0.13,1.615) rectangle (0.13,1.885);
\draw[fill=white, draw=black!35, rounded corners=2pt]
    (-0.5,1.06) rectangle (0.5,1.44);
\fill[blue!60] (0,1.25) circle (0.10cm);
\draw[fill=white, draw=black!35, rounded corners=2pt]
    (-0.5,0.56) rectangle (0.5,0.94);
\fill[red!65] (-0.09,0.66) rectangle (0.09,0.84);
\draw[fill=white, draw=black!35, rounded corners=2pt]
    (-0.5,0.06) rectangle (0.5,0.44);
\fill[red!65] (0,0.25) circle (0.155cm);
\node[font=\scriptsize\itshape, text=gray] at (0,-0.25) {observes};
\draw[arr] (1.15,1.25) --
    node[above, font=\small] {$m \in V^L$}
    node[below, font=\scriptsize, gray] {message}
    (3.05,1.25);
\draw[rounded corners=5pt, fill=gray!7, draw=black!50]
    (3.05,-0.5) rectangle (5.35,2.7);
\node[font=\small\bfseries, align=center] at (4.2,3.1)
    {Agent B \\ {\normalfont\scriptsize(receiver)}};
\draw[fill=orange!25, draw=black!50, rounded corners=2pt]
    (3.7,2.11) rectangle (4.7,2.49);
\node[font=\normalsize\bfseries] at (4.2,2.30) {?};
\draw[fill=white, draw=black!35, rounded corners=2pt]
    (3.7,1.56) rectangle (4.7,1.94);
\fill[blue!60] (4.07,1.615) rectangle (4.33,1.885);
\draw[fill=white, draw=black!35, rounded corners=2pt]
    (3.7,1.06) rectangle (4.7,1.44);
\fill[blue!60] (4.2,1.25) circle (0.10cm);
\draw[fill=white, draw=black!35, rounded corners=2pt]
    (3.7,0.56) rectangle (4.7,0.94);
\fill[red!65] (4.11,0.66) rectangle (4.29,0.84);
\draw[fill=white, draw=black!35, rounded corners=2pt]
    (3.7,0.06) rectangle (4.7,0.44);
\fill[red!65] (4.2,0.25) circle (0.155cm);
\node[font=\scriptsize\itshape, text=gray] at (4.2,-0.25) {observes + selects};
\draw[darr]
    (4.2,-0.5) -- (4.2,-1.0)
    -- node[below, font=\scriptsize]
       {feedback: outcome $+$ true target (both agents)}
    (0,-1.0) -- (0,-0.5);
\end{tikzpicture}
\caption{The referential signaling game.
Each round, Agent~A (sender) observes four candidate objects (white
slots) and a designated target (orange), then emits a fixed-length
symbolic message $m$ from vocabulary $V$.
Agent~B (receiver) observes the same four candidates and the message,
then guesses the target (orange, initially unknown).
Both agents receive full feedback after each round.
Objects (pool of 8) have three binary features: color (red/blue), shape
(circle/square), size (small/large).
Chance accuracy is $0.25$.
Unlike the minimal Lewis game, the receiver always chooses from four
candidates rather than the full object space, creating constant
discrimination pressure.}
\label{fig:game}
\end{figure}
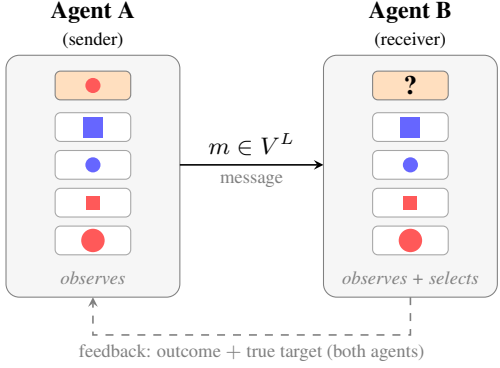

\paragraph{The Game}
Two agents interact for $N = 200$ rounds.
Agent A (the sender) observes four candidate objects sampled uniformly from a pool of eight ($\{\text{red, blue}\} \times \{\text{circle, square}\} \times \{\text{small, large}\}$) and one designated target.
Agent A then emits a symbolic message of fixed length $L$, drawn from a constrained vocabulary $V$.
Agent B (the receiver) observes the same four candidates and the message, then guesses the target.
After each round, both agents observe the outcome (correct or incorrect) and the true target.
Communication is strictly one-directional, and agents never see each other's private memory.
After each round, memory updates use the true target revealed in the feedback.
Chance accuracy is 0.25, since the receiver chooses among four candidates.
Figure~\ref{fig:game} illustrates the game structure, and Figure~\ref{fig:prompts} shows prompt templates for both agents.
No semantics are pre-assigned, so conventions must emerge through play.

\paragraph{Memory Architectures}
We compare five conditions.
In all of them, each agent receives a rolling window of its last 20 \texttt{(message, target, success)} interactions as context each round.
We hold this window fixed at 20 while comparing memory architectures and sweeping capacity, instead of exposing the model's full context, so that performance differences reflect the persistent store each architecture adds rather than the amount of raw history shown.
Study~3 then varies the window size directly ($m \in \{5, 10, 20, 40\}$) to confirm the fixed window is not itself responsible for the results.
The conditions differ in what they add on top; Table~\ref{tab:conditions} summarizes all five.

\begin{table}[t]
\centering
\small
\caption{Memory architecture conditions. All five share a rolling
window of the last 20 interactions as a base. \emph{Update}: how the
persistent store changes each round (overwrite $=$ rewritten in full;
in-place $=$ slot edits; env $=$ compiled by the environment).}
\label{tab:conditions}
\resizebox{\columnwidth}{!}{%
\begin{tabular}{lcccl}
\toprule
Condition & Persistent & Private & Update & Format \\
\midrule
\textbf{memory\_only}   & $\times$   & ---        & ---       & window only \\
\textbf{env\_board}     & \checkmark & $\times$   & env       & convention table \\
\textbf{scratchpad}     & \checkmark & \checkmark & overwrite & free-form notebook \\
\textbf{codebook}       & \checkmark & \checkmark & in-place  & slot list \\
\textbf{codebook\_meta} & \checkmark & \checkmark & in-place  & slot list $+$ meta-note \\
\bottomrule
\end{tabular}}
\end{table}

Every memory store is a field of the agent's structured JSON output (strict \texttt{json\_schema}): the model writes it, the harness parses it and re-injects it into the next round's prompt, and agents never call external tools or edit files directly.
On top of the shared rolling window of the last 20 \texttt{(message, target, success)} triples, each agent also emits a \texttt{rationale} ($\leq 20$ words).
This is logged for analysis only and is neither stored in memory nor transmitted, so the sole signal the receiver ever receives from the sender is the message $m \in V^L$.
The persistent stores differ in how they update.
The scratchpad \texttt{notebook} ($\leq 150$ words) is \emph{overwritten} each round: the agent re-emits the whole notebook and only the latest version is carried forward, so its size does not grow with round count.
The codebook is a fixed-capacity slot list (10 slots) edited \emph{in place} by one structured operation per round (\texttt{append}, \texttt{edit}, or \texttt{none}), and entries persist verbatim until explicitly overwritten.
The \texttt{codebook\_meta} condition adds a single persistent meta-note string, updated the same way after a short warm-up.
Only env\_board is shared: a public convention table the environment compiles from aggregate successful-round counts, which both agents read but neither edits.
Each private store is visible only to its owning agent, and the two agents never see each other's memory.

\paragraph{Channel Configurations}
Capacity $= |V|^L$.
We sweep $|V| \in \{2,3,4,5\}$ and $L \in \{2,3\}$, yielding capacities $\{4, 8, 9, 16, 25, 27, 64, 125\}$.
We use $|V|^L$ as the capacity measure, although Shannon capacity in bits is $\log_2(|V|^L)$, which is monotonically related and yields the same ordering across conditions.

\begin{figure*}[t]
\small
\begin{mdframed}[linewidth=0.6pt,innerleftmargin=8pt,innerrightmargin=8pt,
                 innertopmargin=6pt,innerbottommargin=6pt]

\textbf{Model:} \texttt{gpt-5.4-mini} \quad
\textbf{Temperature:} 1.0 (API default) \quad
\textbf{Response format:} \texttt{json\_schema} (strict) \\[2pt]
\textbf{Rounds:} 200 \quad
\textbf{Candidates per round:} 4

\medskip
\textbf{Both agents (common preamble):}\\[4pt]
\textit{GAME RULES:} Each round: (1) the sender observes 4 candidate objects
and the designated target; (2) the sender emits a fixed-length symbolic
message from the allowed vocabulary; (3) the receiver observes the 4 candidates
and the message, then guesses the target; (4) both agents observe the outcome
(correct/incorrect and the true target).\\[6pt]
\textit{OBJECTS:} Each object has 3 features: color (red/blue), shape
(circle/square), size (small/large). The 8 possible objects are:
\texttt{red\_circle\_small}, \texttt{red\_circle\_large},
\texttt{red\_square\_small}, \texttt{red\_square\_large},
\texttt{blue\_circle\_small}, \texttt{blue\_circle\_large},
\texttt{blue\_square\_small}, \texttt{blue\_square\_large}.

\medskip\hrule\medskip

\begin{minipage}[t]{0.475\textwidth}
\textbf{Agent A (Sender)}\\[3pt]
\textit{CHANNEL:} Allowed vocabulary: \textsl{[A, B, \ldots\ per condition]}.
Messages must be exactly \textsl{[$L$]} tokens. No natural language.\\[3pt]
\textit{STRATEGY:} Develop consistent signal-to-object conventions.
Reuse the same code for the same object type across rounds.
Different objects should receive distinct codes.
Use your interaction memory to track which conventions succeed or fail.
\textsl{[Memory-mode-specific notebook instructions.]}\\[3pt]
\textit{OUTPUT SCHEMA (memory\_only):}
\begin{verbatim}
{ "tokens":   ["X", "Y", ...],
  "rationale": "<= 20 words" }
\end{verbatim}
\textsl{[Scratchpad adds \texttt{notebook} (free text, $\leq$150 words).
Codebook adds \texttt{action} $\in$ \{append, edit, none\},
\texttt{slot} $\in$ 0--9, \texttt{value} (text).
Codebook\_meta also adds \texttt{meta\_note} (single-line text).]}\\[3pt]
\textit{Output sample (round 7):}
\begin{verbatim}
{ "tokens": ["A","B","A"],
  "rationale": "A B A maps to
   red_circle_small; target
   blue_circle_small identified
   by elimination." }
\end{verbatim}
\end{minipage}
\hfill
\begin{minipage}[t]{0.475\textwidth}
\textbf{Agent B (Receiver)}\\[3pt]
\textit{CHANNEL:} Agent A's messages use vocabulary:
\textsl{[A, B, \ldots\ per condition]}.
Messages contain exactly \textsl{[$L$]} tokens.\\[3pt]
\textit{STRATEGY:} Learn Agent A's signal conventions from interaction
history. Track which messages map to which objects.
Prefer interpretations most consistent with past successful rounds.
\textsl{[Memory-mode-specific notebook instructions.]}\\[3pt]
\textit{OUTPUT SCHEMA (memory\_only):}
\begin{verbatim}
{ "choice":    <int, 1-4 = candidate index>,
  "rationale": "<= 20 words" }
\end{verbatim}
\textit{Output sample (round 7):}
\begin{verbatim}
{ "choice": 1,
  "rationale": "A B A best matches
   blue_circle_small from prior
   patterns; it shares blue,
   circle, and small." }
\end{verbatim}
\end{minipage}

\end{mdframed}
    \caption{Prompt template and output schema for both agents.
The common preamble (top) is identical in both system prompts.
Agent-specific sections supply role, channel constraints, strategy,
and structured output schema. Italic placeholders are filled per
condition; no object meanings are pre-assigned.}
\label{fig:prompts}
\end{figure*}

\paragraph{Metrics}
Let $\mathcal{O}$ be the set of 8 objects, $W$ a set of rounds, $o_t$ the target and $\hat{o}_t$ the receiver's guess at round $t$, $m_t \in V^L$ the sender's message, and $\mathcal{F} = \{\text{color, shape, size}\}$ the three binary features.
All language metrics are computed over the late-game window $W = \{151,\ldots,200\}$ unless stated otherwise, since the early game contains exploration and unstable conventions and the late window best reflects the code the agents have settled on.

\medskip\noindent
\textbf{Accuracy} (reported in 50-round windows) measures whether the agents actually coordinate on the task:
\[
\mathrm{Acc}(W) \;=\; \frac{1}{|W|}\sum_{t\in W}\mathbf{1}[\hat{o}_t = o_t].
\]

\noindent \textbf{TopSim} \citep{brighton2006understanding} measures how far the geometry of meaning space agrees with that of signal space (topographic structure), making it our main indicator of compositional structure.
To apply it to noisy multi-round play, we estimate a sender \emph{effective codebook} $c:\mathcal{O}\to V^L$ by taking the modal message for each object over $W$: $c(o) = \operatorname*{arg\,max}_{m}\,\#\{t\!
\in\!
W : o_t\!=\!o,\, m_t\!=\!m\}$.
Let $d_S(o,o') = \sum_{k=1}^{3}\mathbf{1}[f_k(o)\neq f_k(o')]$ be the feature-Hamming semantic distance ($0$--$3$) and $d_H(o,o') = \sum_{l=1}^{L}\mathbf{1}[c(o)_l\neq c(o')_l]$ the message Hamming distance.
\[
\mathrm{TopSim}(W) \;=\; \rho_S\!
\Bigl(\bigl\{d_S(o_i,o_j)\bigr\}_{i<j},\, \bigl\{d_H(o_i,o_j)\bigr\}_{i<j}\Bigr),
\]
where $\rho_S$ is Spearman rank correlation over all $\binom{8}{2}=28$ object pairs, with $+1$ indicating perfect compositionality.

\noindent \textbf{Best MI}: for token position $p\in\{1,\ldots,L\}$ and feature $k\in\{1,\ldots,|\mathcal{F}|\}$, estimate $I(P_p;\,F_k)$ empirically from the per-round pairs $\{(m_t[p],\,f_k(o_t))\}_{t\in W}$.
This captures positional slot structure even when the full codebook is still noisy or only partially compositional:
\[
\mathrm{MI}^*(W) \;=\; \max_{p,\,k}\; I(P_p;\,F_k).
\]

\noindent \textbf{Collision rate} measures ambiguity in the induced lexicon by checking how often distinct objects collapse onto the same effective message.
Using the same effective codebook $c$,
\[
\mathrm{Coll}(W) \;=\; \frac{\bigl|\{o\in\mathcal{O} : \exists\,o'\neq o,\; c(o)=c(o')\}\bigr|}{|\mathcal{O}|}.
\]

\begin{figure*}[t]
    \centering
    \includegraphics[width=0.75\textwidth]{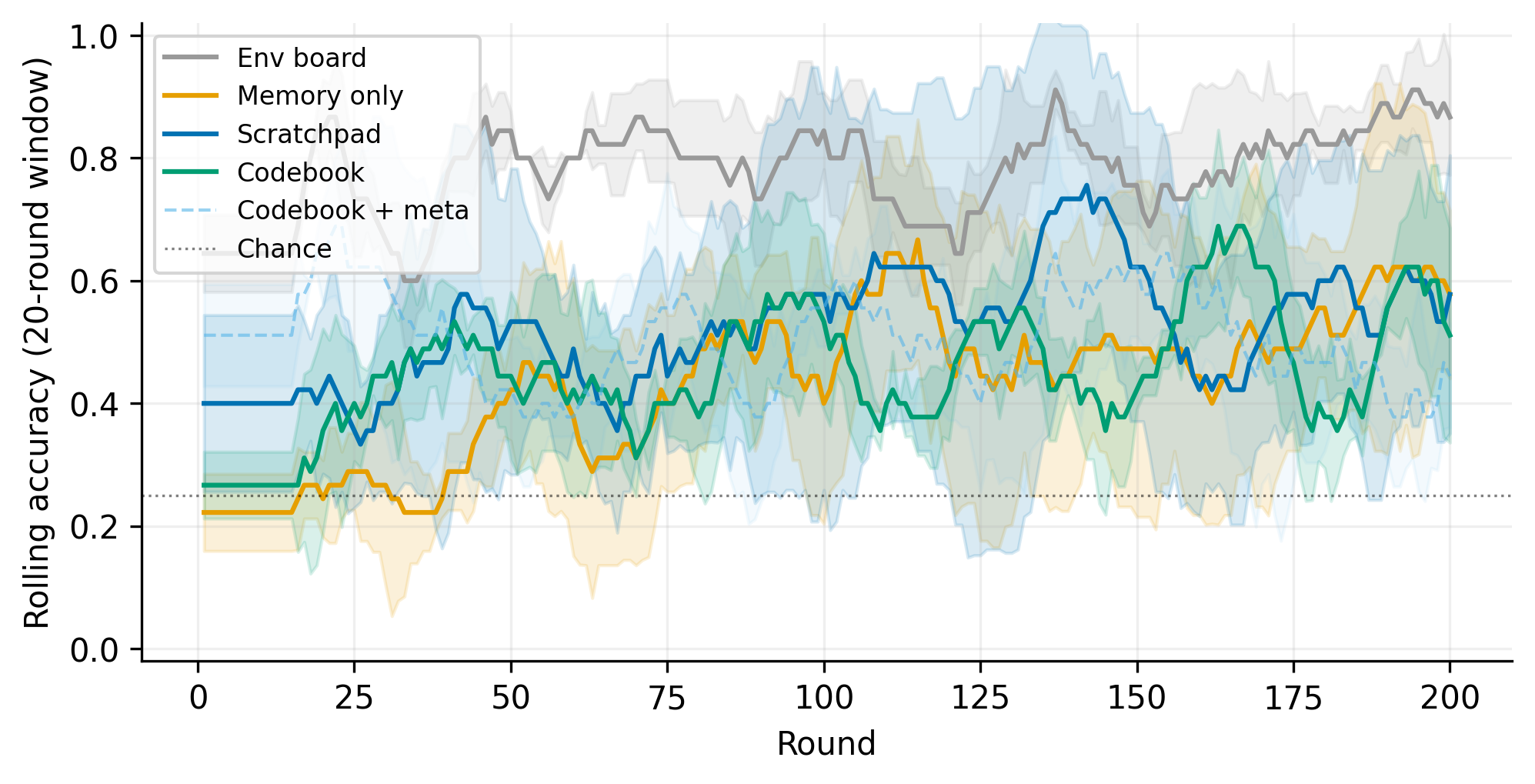}
    \vspace{-8pt}
    \caption{
        Learning dynamics across memory architectures (cap $= 27$,
        3 seeds, 15-round rolling mean $\pm$ std).
        Env\_board converges quickly via the shared public table rather
        than forming conventions.
        Scratchpad shows the steepest mid-game rise but drops in late
        rounds with widening cross-seed variance.
        Memory\_only is the most stable late-game.
        Codebook modes show high variance throughout with no sustained
        improvement. Dotted line: chance ($0.25$).
    }
    \label{fig:fig2_learning_curves}
\end{figure*}

\section{Study 1: Memory Architecture at Fixed Capacity}
\label{sec:study1}

We fix the channel at $|V| = 3$, $L = 3$ (cap $= 27$) and compare all five memory architectures across 200 rounds, replicated over three random seeds $\{7, 42, 123\}$.
Tables~\ref{tab:memory_accuracy} and~\ref{tab:memory_language} report windowed accuracy and late-game language metrics respectively; Figure~\ref{fig:fig2_learning_curves} shows the learning dynamics.

{\setlength{\intextsep}{1pt}\setlength{\abovecaptionskip}{2pt}\setlength{\belowcaptionskip}{0pt}
\begin{table}[t]
    \centering
    \caption{
        Windowed accuracy across 200 rounds (mean $\pm$ std, 3 seeds).
        Chance $= 0.25$.
    }
    \label{tab:memory_accuracy}
    \renewcommand{\arraystretch}{0.9}
    \resizebox{\columnwidth}{!}{%
    \begin{tabular}{lcccc}
        \toprule
        Mode & R1--50 & R51--100 & R101--150 & R151--200 \\
        \midrule
        env\_board     & .720{\scriptsize$\pm$.05} & .813{\scriptsize$\pm$.06} & .760{\scriptsize$\pm$.02} & \textbf{.827}{\scriptsize$\pm$.09} \\
        scratchpad     & .553{\scriptsize$\pm$.08} & .587{\scriptsize$\pm$.12} & \textbf{.767}{\scriptsize$\pm$.11} & .653{\scriptsize$\pm$.10} \\
        memory\_only   & .400{\scriptsize$\pm$.13} & .547{\scriptsize$\pm$.04} & .640{\scriptsize$\pm$.11} & .660{\scriptsize$\pm$.02} \\
        codebook\_meta & .500{\scriptsize$\pm$.13} & .493{\scriptsize$\pm$.11} & .540{\scriptsize$\pm$.16} & .460{\scriptsize$\pm$.14} \\
        codebook       & .393{\scriptsize$\pm$.01} & .473{\scriptsize$\pm$.10} & .433{\scriptsize$\pm$.04} & .527{\scriptsize$\pm$.13} \\
        \bottomrule
    \end{tabular}}
\end{table}}
{\setlength{\intextsep}{1pt}\setlength{\abovecaptionskip}{2pt}\setlength{\belowcaptionskip}{0pt}
\begin{table}[t]
    \centering
    \caption{
        Late-game language metrics, R151--200
        (mean $\pm$ std, 3 seeds).
    }
    \label{tab:memory_language}
    \renewcommand{\arraystretch}{0.9}
    \resizebox{\columnwidth}{!}{%
    \begin{tabular}{lcccc}
        \toprule
        Mode & Accuracy & TopSim & Best MI & Collision \\
        \midrule
        env\_board     & \textbf{.827}{\scriptsize$\pm$.090} & $-.002${\scriptsize$\pm$.223} & .527{\scriptsize$\pm$.118} & .250{\scriptsize$\pm$.250} \\
        memory\_only   & .660{\scriptsize$\pm$\textbf{.020}} & \textbf{.347}{\scriptsize$\pm$.232} & \textbf{.939}{\scriptsize$\pm$.073} & .750{\scriptsize$\pm$.250} \\
        scratchpad     & .653{\scriptsize$\pm$.095} & .000{\scriptsize$\pm$.147} & .451{\scriptsize$\pm$.109} & \textbf{.250}{\scriptsize$\pm$.250} \\
        codebook       & .527{\scriptsize$\pm$.129} & .179{\scriptsize$\pm$.272} & .359{\scriptsize$\pm$.271} & .625{\scriptsize$\pm$.217} \\
        codebook\_meta & .460{\scriptsize$\pm$.144} & .116{\scriptsize$\pm$.064} & .322{\scriptsize$\pm$.090} & .458{\scriptsize$\pm$.191} \\
        \bottomrule
    \end{tabular}}
\end{table}}

These results separate performance from language quality.
The env\_board condition achieves the highest late-game accuracy ($0.827 \pm 0.09$) but produces near-zero TopSim, indicating a memorized lookup table rather than a productive code.
Because meaning is read off a shared public board, no internal convention needs to form, so we exclude env\_board from compositionality analysis.

Within the private-memory conditions, scratchpad shows partial positional structure.
It peaks at R101--150 ($0.767 \pm 0.11$), and in each seed at least one token position comes to encode a single feature.
Which feature lands on which position varies across seeds, however, and no seed at this capacity settles on a clean global code, so the late-game TopSim in Table~\ref{tab:memory_language} stays near zero.
A fuller positional code appears in some higher-capacity runs (Study~2), where the sender factors color and shape onto separate positions.

By contrast, memory\_only is reliable without being fully productive.
It is the most stable mode across seeds (late-game std $= 0.020$, the lowest in the study) and achieves the highest late-game TopSim and mutual information.
Yet the 75\% collision rate tells a different story: the global code remains overlapping, and the receiver exploits the four-candidate context to resolve ambiguity locally rather than decoding a clean global convention.

The slot-based conditions show the opposite pattern: fast early organization without durable consolidation.
Codebook reaches peak compositional structure earliest (P1$\to$color purity $= 0.96$ at 50 rounds) but ends at only $0.527 \pm 0.13$ late-game accuracy.
The slot-based format bootstraps early convergence, but with no pruning mechanism, stale and conflicting entries accumulate over 200 rounds and pull performance down.
Codebook\_meta, which pairs the slot list with a persistent abstract meta-note for language-level rules, ends lower still ($0.460 \pm 0.14$), making it the weakest private-memory condition.
Inspecting the actual meta-notes across runs, however, reveals why: rather than developing structural insights such as ``position 1 encodes color,'' the note freezes by round 30 into generic operational reminders along the lines of \textit{``3-token fixed code; reuse confirmed mappings; edit only on failure''} and stays there for the rest of the run.
The meta-note never becomes higher-order; it simply restates what the agent prompt already instructs.
Meanwhile, the slot list accumulates conflicts of its own.
The result is an agent with two sources of guidance that can contradict each other and no mechanism for resolving the tension, which is worse than the simpler codebook it extends.

\begin{figure*}[t]
    \centering
    \includegraphics[width=0.72\textwidth]{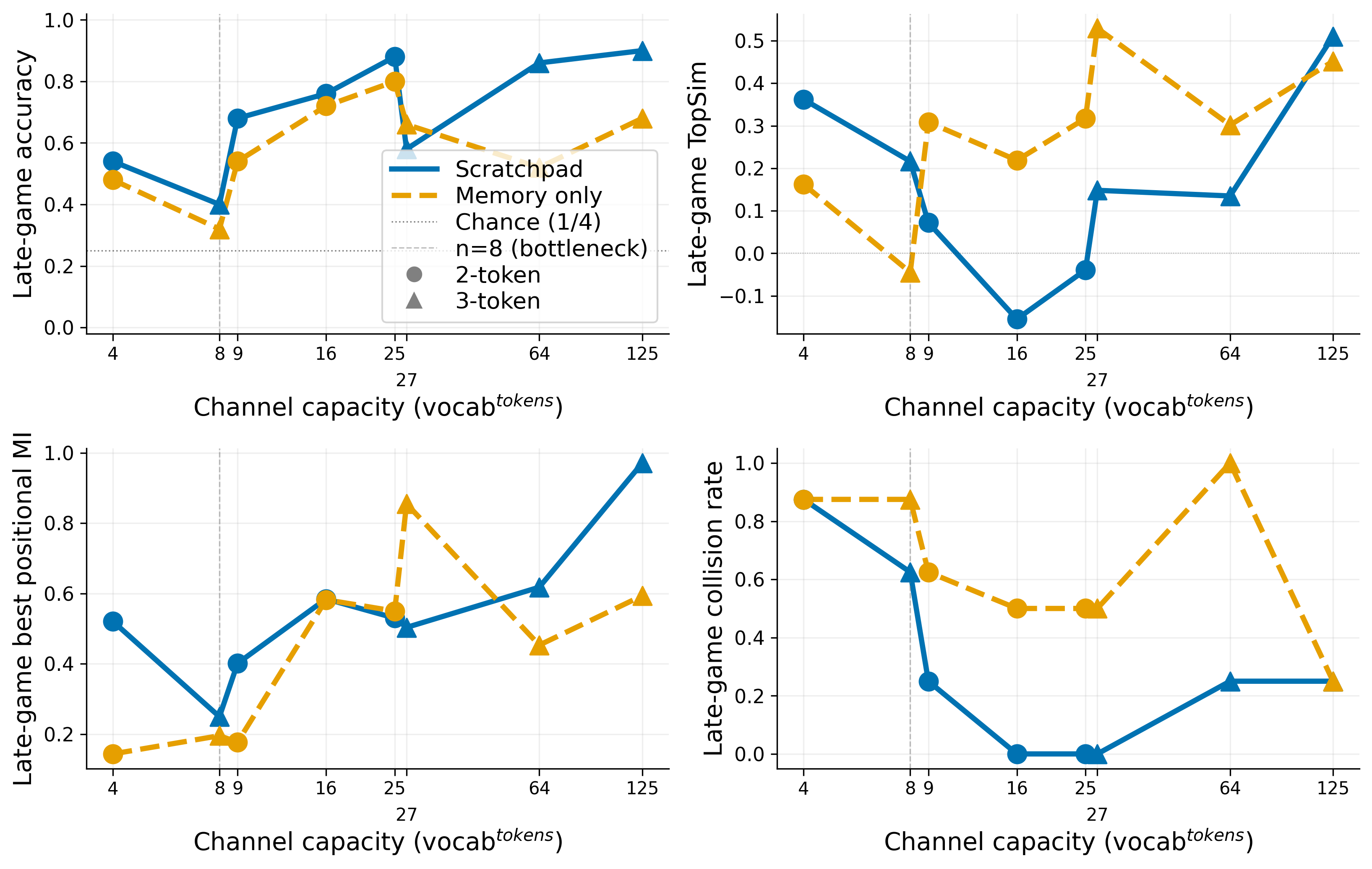}
    \caption{
        Capacity curves for scratchpad and memory\_only (R151--200,
        seed $= 7$). Circles: 2-token configs; triangles: 3-token.
        Dashed line: predicted bottleneck (cap $= 8$).
        \textbf{(a)}~Scratchpad accuracy rises with capacity within each
        token-length family; memory\_only peaks at cap $= 25$ then collapses.
        \textbf{(b)}~TopSim does not track accuracy: memory\_only peaks
        earlier and higher in TopSim despite lower accuracy.
        \textbf{(c)}~MI follows a similar divergence, with scratchpad
        rising steadily and memory\_only peaking mid-range then declining.
        \textbf{(d)}~Collision rate separates the two modes most clearly:
        scratchpad reaches zero at moderate capacity; memory\_only hits
        $1.0$ at cap $= 64$.
    }
    \label{fig:fig3_capacity_curves}
\end{figure*}

\section{Study 2: Capacity and Language Quality}
\label{sec:study2}

Study~1 established three things.
First, env\_board achieves high accuracy by reading from a shared table rather than forming conventions, so it tells us little about private convention formation.
Second, the codebook conditions produced the weakest and most inconsistent results across seeds.
Third, scratchpad and memory\_only showed meaningfully different learning trajectories and compositionality profiles.
Study~2 therefore focuses on these two architectures and asks how their behavior changes as we vary channel capacity from 4 to 125, excluding env\_board, codebook, and codebook\_meta.

We sweep all 16 channel configurations (8 capacities $\times$ 2 modes: scratchpad and memory\_only) for 200 rounds at seed $= 7$ (Table~\ref{tab:capacity_sweep}), then replicate three key conditions across multiple seeds (Table~\ref{tab:capacity_replication}).
Cap~$= 8$ was replicated with 8 seeds because the high initial variance of three seeds made them unreliable as a characterization of the distribution.

{\setlength{\intextsep}{2pt}
\begin{table}[t]
    \centering
    \caption{
        Rate-distortion sweep: late-game accuracy (R151--200),
        seed $= 7$. Capacity $= |V|^L$.
        Bold: best per column per mode.
    }
    \label{tab:capacity_sweep}
    \small
    \begin{tabular}{llrcccc}
        \toprule
        Mode & Config & Cap & Acc & TopSim & Best MI & Coll. \\
        \midrule
        \multirow{8}{*}{\rotatebox{90}{scratchpad}}
         & v2\_t2 &   4 & 0.54 &  0.362 & 0.521 & 0.875 \\
         & v2\_t3 &   8 & 0.40 &  0.216 & 0.249 & 0.625 \\
         & v3\_t2 &   9 & 0.68 &  0.072 & 0.401 & 0.250 \\
         & v4\_t2 &  16 & 0.76 & $-.155$ & 0.584 & \textbf{0.000} \\
         & v5\_t2 &  25 & 0.88 & $-.039$ & 0.530 & \textbf{0.000} \\
         & v3\_t3 &  27 & 0.58 &  0.148 & 0.503 & \textbf{0.000} \\
         & v4\_t3 &  64 & 0.86 &  0.135 & 0.618 & 0.250 \\
         & v5\_t3 & 125 & \textbf{0.90} & \textbf{0.510} & \textbf{0.971} & 0.250 \\
        \midrule
        \multirow{8}{*}{\rotatebox{90}{memory\_only}}
         & v2\_t2 &   4 & 0.48 &  0.162 & 0.144 & 0.875 \\
         & v2\_t3 &   8 & 0.32 & $-.045$ & 0.195 & 0.875 \\
         & v3\_t2 &   9 & 0.54 &  0.308 & 0.177 & 0.625 \\
         & v4\_t2 &  16 & 0.72 &  0.219 & 0.582 & 0.500 \\
         & v5\_t2 &  25 & \textbf{0.80} & 0.317 & 0.550 & 0.500 \\
         & v3\_t3 &  27 & 0.66 & \textbf{0.529} & \textbf{0.855} & 0.500 \\
         & v4\_t3 &  64 & 0.52 &  0.302 & 0.453 & \textbf{1.000} \\
         & v5\_t3 & 125 & 0.68 &  0.452 & 0.594 & 0.250 \\
        \bottomrule
    \end{tabular}
\end{table}}
{\setlength{\intextsep}{2pt}
\begin{table}[t]
    \centering
    \caption{
        Study~2 replication: key conditions (mean $\pm$ std, R151--200).
        Cap~$= 8$: $n = 8$ seeds (1--5, 7, 42, 123);
        caps~25 and~64: $n = 3$ seeds (7, 42, 123).
    }
    \label{tab:capacity_replication}
    \resizebox{\columnwidth}{!}{%
    \begin{tabular}{llcccc}
        \toprule
        Mode & Cap & Acc & TopSim & Best MI & Coll. \\
        \midrule
        scratchpad   & \phantom{0}8 & .542{\scriptsize$\pm$.166} & .161{\scriptsize$\pm$.179} & .434{\scriptsize$\pm$.265} & .609{\scriptsize$\pm$.216} \\
        scratchpad   & 25           & \textbf{.867}{\scriptsize$\pm$.023} & .099{\scriptsize$\pm$.266} & .769{\scriptsize$\pm$.227} & .333{\scriptsize$\pm$.382} \\
        scratchpad   & 64           & .760{\scriptsize$\pm$.092} & .264{\scriptsize$\pm$.139} & .683{\scriptsize$\pm$.145} & .167{\scriptsize$\pm$.144} \\
        \midrule
        memory\_only & \phantom{0}8 & .542{\scriptsize$\pm$.151} & .201{\scriptsize$\pm$.200} & .502{\scriptsize$\pm$.257} & .797{\scriptsize$\pm$.148} \\
        memory\_only & 25           & .747{\scriptsize$\pm$.076} & \textbf{.383}{\scriptsize$\pm$.137} & .669{\scriptsize$\pm$.274} & .625{\scriptsize$\pm$.125} \\
        memory\_only & 64           & .580{\scriptsize$\pm$.140} & .254{\scriptsize$\pm$.046} & .554{\scriptsize$\pm$.148} & .833{\scriptsize$\pm$.191} \\
        \bottomrule
    \end{tabular}}
\end{table}}

\paragraph{Two architectures, two capacity curves.}
Figure~\ref{fig:fig3_capacity_curves} shows the two modes diverging as capacity grows.
Scratchpad accuracy increases with capacity within each token-length family: $0.54 \to 0.88$ across 2-token configurations and $0.40 \to 0.90$ across 3-token configurations.
Scratchpad collision drops to zero at cap $= 16$--$27$, but rises back to $0.25$ at cap $= 64$ and $125$; persistent notes enable stable global codes at moderate capacity, though some ambiguity re-emerges at the largest signal spaces.
Memory\_only follows a different path entirely.
Accuracy peaks at cap $= 25$ ($0.80$), falls to $0.52$ at cap $= 64$ where collision hits $1.0$: every object's most-common message coincides with another object's.
At cap $= 125$, memory\_only partially recovers to $0.68$, but scratchpad reaches $0.90$ there, so the gap widens.
Without a persistent note, the 20-round window cannot accumulate enough evidence per code as the space expands.
Memory\_only achieves its highest MI ($0.855$) at cap $= 27$, above scratchpad's MI ($0.503$) at that capacity.
This repeats the Study~1 pattern where high MI co-occurs with high collision, indicating locally structured but globally ambiguous codes rather than a productive language.

\paragraph{Token length matters independently of capacity.}
Across both modes, 3-token configurations consistently underperform 2-token configurations at comparable capacity levels.
Message length appears to interact with convergence difficulty beyond what raw capacity captures: longer messages mean more positions to coordinate, even when the total signal space is equivalent.

\paragraph{At the bottleneck, outcomes are bimodal.}
Cap $= 8$ showed the widest variance of any condition, so we ran eight seeds in total to characterize the distribution.
The result is not a noisy average but a split: runs either succeed (accuracy $\geq 0.66$) or plateau well below the performance seen at higher capacities ($\leq 0.56$), with little in between.
An exact permutation test shows cap $= 8$ is significantly below the higher-capacity cap $= 25$ condition ($p = 0.024$, scratchpad).
With eight objects and eight signals, channel capacity equals source entropy ($\log_2 8 = 3$ bits on both sides), so the codebook must be perfectly injective: every object needs a distinct signal, and there are no spares.
Above the bottleneck, surplus signals let agents repair an early collision by reassigning an object to an unused code.
At the bottleneck there is no surplus beyond the eight signals a perfect code needs, so repair requires both agents to discover and agree on a signal the collision leaves idle, which they rarely coordinate.
A collision that forms in the first few rounds usually persists, and the run stays low for the remaining 190 rounds.
Which outcome occurs appears to depend strongly on the object sequence in the early rounds.
Both scratchpad and memory\_only land at the same mean ($0.542$), the only condition in this study where the persistent-note advantage disappears entirely.

\paragraph{Surplus capacity stabilizes conventions.}
At cap $= 25$, scratchpad achieves $0.867 \pm 0.023$, the tightest result of any multi-seed condition in this study.
Memory\_only also replicates cleanly ($0.747 \pm 0.076$) at this sweet spot: enough signal space to absorb early errors without overwhelming the window.
The memory\_only collapse at cap $= 64$ recurs in two of three seeds ($0.580 \pm 0.140$, collision $0.833 \pm 0.191$) and points to code proliferation, though at $n = 3$ the drop is not statistically distinguishable from sampling noise.

\section{Study 3: Consolidation vs.\ History Length}
\label{sec:study3}

Study~2 identified a clear performance divergence.
Scratchpad continues to perform well at high capacity, while memory\_only peaks and then collapses.
This collapse could have two explanations.
The straightforward explanation is that the 20-round window is too short to track a large code space, so extending it would close the gap.
The structural explanation is that stateless agents rely on the rolling window as their only carrier of conventions.
Under this account, any convention that falls out of the most recent $m$ rounds is simply lost, regardless of how long the window is.
We distinguish these explanations by sweeping the memory window size $m \in \{5, 10, 20, 40\}$ at cap $= 64$ and cap $= 25$, holding all other parameters fixed.
Only scratchpad and memory\_only are tested here, for the same reason as in Study~2.

\begin{figure}[t]
    \centering
    \includegraphics[width=\columnwidth]{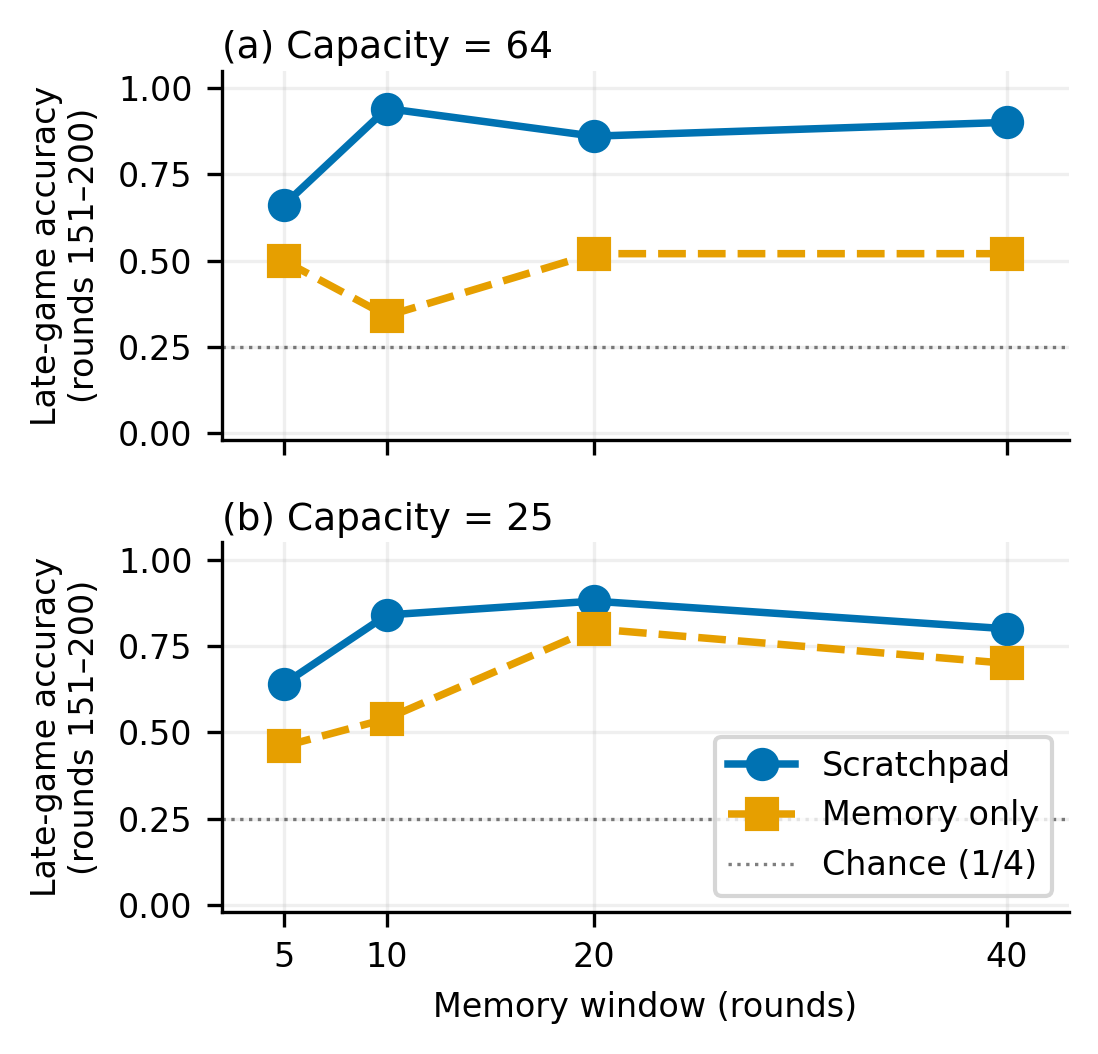}
    \vspace{-18pt}
    \caption{
        Effect of memory window size on late-game accuracy (R151--200,
        seed $= 7$).
        \textbf{(a)}~Capacity $= 64$: scratchpad reaches $0.94$ with
        $m = 10$ rounds; memory\_only stays low across all window sizes.
        \textbf{(b)}~Capacity $= 25$: both architectures peak near
        $m = 20$ and dip at $m = 40$.
    }
    \label{fig:fig4_memory_cap}
\end{figure}

Figure~\ref{fig:fig4_memory_cap} shows results.
At cap $= 64$, memory\_only accuracy stays low at every window size: $0.50$, $0.34$, $0.52$, $0.52$ for $m = 5, 10, 20, 40$.
Doubling the window from 20 to 40 rounds produces no improvement.
Scratchpad reaches $0.94$ with only $m = 10$ rounds of context, which is less history than memory\_only ever uses.
At cap $= 25$, however, both architectures improve up to $m = 20$: memory\_only climbs from $0.46$ to $0.80$ and scratchpad from $0.64$ to $0.88$. At $m = 40$, both modes dip (scratchpad $0.80$, memory\_only $0.70$), suggesting a context-window sweet spot near 20 rounds, as opposed to a simple ``more is better'' relationship.
Thus, rolling-window memory can work when the code space is manageable, and its failure is specific to high capacity.

\vspace{-8pt}

\section{Discussion}

\paragraph{Connections to Information Theory.}
Shannon's capacity theorem \citep{shannon1948mathematical} sets a hard floor on a context-independent object code: any channel carrying fewer than $\log_2|\mathcal{O}| = 3$ bits cannot distinguish all eight objects, regardless of encoding strategy.
The information bottleneck (IB) framework \citep{tishby1999information} characterizes optimal \emph{rate--relevance} trade-offs: how compressed a representation can be while preserving information about a target variable.
In emergent neural communication, \citet{resnick2020capacity} relate compositionality to both channel bandwidth and model capacity and posit an intermediate \emph{range} of settings rather than a unique peak.
A common IB-inspired heuristic for referential games is nonetheless to scrutinize tight channels whose capacity is on the order of the number of referents, including $|V|^L = 8$ when $|\mathcal{O}| = 8$.
That point is not a compositional optimum here.
Capacity matters, but cap $= 8$ is a fragility point.
With no redundancy, early misalignment is rarely repaired, making outcomes run-dependent.
Surplus capacity is generally better, echoing how natural lexicons sit near but not on the IB efficiency frontier, keeping some redundancy \citep{zaslavsky2018efficient,regier2015word}.
Idealized bottleneck analyses treat agents as if they could reach the optimal rate--relevance frontier.
LLM agents are not optimal encoders: they negotiate conventions through interaction, and how close they get to any such frontier depends on whether they can consolidate what they learn.
Memory architecture is the variable the framework leaves out.

\paragraph{Why window size does not matter at high capacity.}
Study~3 shows the collapse is a consolidation problem rather than a history problem: scratchpad agents succeed with as few as 10 rounds of context, while doubling the stateless window from 20 to 40 rounds yields no improvement.
This mirrors the distinction \citet{kirby2015compression} draw between communication pressure (expressivity, imposed during interaction) and compression pressure (learnability, imposed during consolidation), and it connects to the working-memory capacity limit \citep{miller1956magical} and to formal accounts of memory compression under a token budget \citep{talebirad2026hierarchical}: as the code space grows, a fixed window provides diminishing evidence for each code, and coordination fails regardless of window size.

\paragraph{Sender and receiver representations.}
Both agents usually converge on holistic per-message lookup tables rather than compositional codes, and the two sides track each other closely.
Across runs, the mutual information a whole message carries about the true target is tightly correlated with the information it carries about the receiver's choice (Pearson $r = 0.87$ late-game, similar over the full game), with the sender's side only slightly higher.
A compositional sender rule that factors features onto token positions appears only occasionally.
The clearest case is a cap~$= 25$ run whose sender notebook read ``\texttt{A=red, B=small, C=blue, D=circle, E=square.
Use color+shape as primary code; size only if needed.}''; its receiver kept a holistic list but still reached $0.88$ accuracy via the four-candidate context.
Sender-lexicon ambiguity limits coordination more than receiver decoding does: most senders map the eight objects onto six or fewer distinct messages, and the receiver decodes nearly as well as the best fixed message-to-object decoder.

\paragraph{Convention drift.}
Inspection of scratchpad sender notebooks suggests \emph{convention drift} as a mechanism for the late-game accuracy dip: rather than committing to established codes, the sender re-assigns the same token sequence to different objects over the run, invalidating the receiver's accumulated experience and causing performance to regress.
This is not a context-length effect: input token counts stay flat near 780 (sender) throughout rounds 25--200, well within any modern LLM's range.

\section{Limitations and Future Work}

All experiments use a single model (\texttt{gpt-5.4-mini}), fixed sender/receiver roles, an eight-object space, and a four-way discrimination task each round.
Most capacity conditions run at a single seed.
We replicate three key conditions across seeds, with cap~$= 8$ at $n = 8$, and the memory window sweep uses a single seed.
Given non-reproducible LLM outputs and $n \leq 3$ for the replicated conditions, we report means and sample standard deviations and run significance tests only where the sample size supports them, as at cap~$= 8$.
Results should be read as indicative rather than statistically conclusive, and where cross-seed variance is wide, as in several capacity conditions and Figure~\ref{fig:fig3_capacity_curves}, the differences are trends rather than established effects.
Replicating the sweep on open-weight models, larger compositional spaces, and more seeds per condition would show whether the observed curves are model-specific or reflect general LLM-agent properties, and would put the cross-seed comparisons on firmer statistical footing.

We do not yet test interventions aimed at the failure modes we identify.
Convention drift offers a clear target: instructing the sender to treat established mappings as immutable, or otherwise stabilizing established conventions, may remove the late-game accuracy drop without sacrificing early flexibility.
Acting on the sender's lexicon directly, by rewarding distinct codes or penalizing reuse, would test whether reducing collisions improves coordination more than receiver-side changes do.
A richer sender output schema, with explicit room to plan a positional code rather than only emit tokens and a brief rationale, could test whether compositional encoding emerges more reliably than the partial, run-dependent structure we observe.
Finally, combining within-run accumulation with cross-generational transmission \citep{kouwenhoven2025searching} could test whether iterated learning suppresses drift through transmission pressure.

\section{Conclusion}

Memory architecture strongly shapes whether LLM agents converge on a stable language.
Persistent notebooks let agents consolidate conventions and benefit from surplus channel capacity, while stateless agents degrade once the code space outgrows the rolling window.
Channel capacity matters, but not as bottleneck reasoning predicts.
Cap $= 8$ is a fragility point rather than an optimum, and surplus capacity gives conventions room to stabilize.
At the same time, notebooks introduce their own failure mode when agents revise established mappings mid-game.
Thus, our study shows that signals become language only when channel capacity and memory architecture jointly support stable conventions.
\vspace{8pt}
\section*{Acknowledgments}
This research was supported by the Alberta Machine Intelligence Institute (Amii) and the CIFAR Canada AI Chair program.
We also thank the Network for Applied Technology (NAT) for its support.
The authors used Claude (Anthropic) to assist with code development and manuscript editing.
All AI-assisted outputs were reviewed and verified by the authors, who take full responsibility for the work.

\footnotesize
\bibliographystyle{apalike}
\bibliography{references}

\begin{thebibliography}{}

\bibitem[Akata et~al., 2025]{akata2025playing}
Akata, E., Schulz, L., Coda-Forno, J., Oh, S.~J., Bethge, M., and Schulz, E.
  (2025).
\newblock Playing repeated games with large language models.
\newblock {\em Nature Human Behaviour}, 9(7):1380--1390.

\bibitem[Ashery et~al., 2025]{ashery2025emergent}
Ashery, A.~F., Aiello, L.~M., and Baronchelli, A. (2025).
\newblock Emergent social conventions and collective bias in {LLM} populations.
\newblock {\em Science Advances}, 11(20):eadu9368.

\bibitem[Brighton and Kirby, 2006]{brighton2006understanding}
Brighton, H. and Kirby, S. (2006).
\newblock Understanding linguistic evolution by visualizing the emergence of
  topographic mappings.
\newblock {\em Artificial Life}, 12:229--242.

\bibitem[Brown et~al., 2020]{brown2020language}
Brown, T.~B. et~al. (2020).
\newblock Language models are few-shot learners.
\newblock In {\em Advances in Neural Information Processing Systems}.

\bibitem[Kirby, 2001]{kirby2001spontaneous}
Kirby, S. (2001).
\newblock Spontaneous evolution of linguistic structure: An iterated learning
  model of the emergence of regularity and irregularity.
\newblock {\em IEEE Transactions on Evolutionary Computation}, 5:102--110.

\bibitem[Kirby et~al., 2008]{kirby2008cumulative}
Kirby, S., Cornish, H., and Smith, K. (2008).
\newblock Cumulative cultural evolution in the laboratory: An experimental
  approach to the origins of structure in human language.
\newblock {\em Proceedings of the National Academy of Sciences},
  105:10681--10686.

\bibitem[Kirby et~al., 2015]{kirby2015compression}
Kirby, S., Tamariz, M., Cornish, H., and Smith, K. (2015).
\newblock Compression and communication in the cultural evolution of linguistic
  structure.
\newblock {\em Cognition}, 141:87--102.

\bibitem[Kouwenhoven et~al., 2025]{kouwenhoven2025searching}
Kouwenhoven, T., Peeperkorn, M., and Verhoef, T. (2025).
\newblock Searching for structure: Investigating emergent communication with
  large language models.
\newblock In {\em International Conference on Computational Linguistics}.

\bibitem[Lazaridou et~al., 2018]{lazaridou2018referential}
Lazaridou, A., Hermann, K.~M., Tuyls, K., and Clark, S. (2018).
\newblock Emergence of linguistic communication from referential games with
  symbolic and pixel input.
\newblock In {\em International Conference on Learning Representations}.

\bibitem[Lazaridou et~al., 2017]{lazaridou2017multi}
Lazaridou, A., Peysakhovich, A., and Baroni, M. (2017).
\newblock Multi-agent cooperation and the emergence of (natural) language.
\newblock In {\em International Conference on Learning Representations}.

\bibitem[Lewis, 1969]{lewis1969convention}
Lewis, D. (1969).
\newblock {\em Convention: A Philosophical Study}.
\newblock Harvard University Press.

\bibitem[Li and Bowling, 2019]{li2019ease}
Li, F. and Bowling, M. (2019).
\newblock Ease-of-teaching and language structure from emergent communication.
\newblock In {\em Advances in Neural Information Processing Systems}.

\bibitem[Lowe et~al., 2019]{lowe2019pitfalls}
Lowe, R., Foerster, J., Boureau, Y., Pineau, J., and Dauphin, Y. (2019).
\newblock On the pitfalls of measuring emergent communication.
\newblock In {\em International Conference on Autonomous Agents and Multi-Agent
  Systems}.

\bibitem[Miller, 1956]{miller1956magical}
Miller, G.~A. (1956).
\newblock The magical number seven, plus or minus two: Some limits on our
  capacity for processing information.
\newblock {\em Psychological Review}, 63:81--97.

\bibitem[Nye et~al., 2021]{nye2021scratchpads}
Nye, M., Andreassen, A.~J., Gur-Ari, G., Michalewski, H., Austin, J., Bieber,
  D., Dohan, D., Lewkowycz, A., Bosma, M., Luan, D., Sutton, C., and Odena, A.
  (2021).
\newblock Show your work: Scratchpads for intermediate computation with
  language models.
\newblock {\em arXiv preprint arXiv:2112.00114}.

\bibitem[Parsaee et~al., 2025]{talebirad2025loopbench}
Parsaee, A., Talebirad, Y., Szepesvári, C., Ohal, V., and Redman, E. (2025).
\newblock {LoopBench}: Discovering emergent symmetry breaking strategies with
  {LLM} swarms.
\newblock {\em arXiv preprint arXiv:2512.13713}.

\bibitem[Regier et~al., 2015]{regier2015word}
Regier, T., Kemp, C., and Kay, P. (2015).
\newblock Word meanings across languages support efficient communication.
\newblock In MacWhinney, B. and O'Grady, W., editors, {\em The Handbook of
  Language Emergence}, pages 237--263. Wiley.

\bibitem[Ren et~al., 2020]{ren2020compositional}
Ren, Y., Guo, S., Labeau, M., Cohen, S.~B., and Kirby, S. (2020).
\newblock Compositional languages emerge in a neural iterated learning model.
\newblock In {\em International Conference on Learning Representations}.

\bibitem[Resnick et~al., 2020]{resnick2020capacity}
Resnick, C., Gupta, A., Foerster, J., Dai, A.~M., and Cho, K. (2020).
\newblock Capacity, bandwidth, and compositionality in emergent language
  learning.
\newblock In {\em International Conference on Autonomous Agents and Multi-Agent
  Systems}.

\bibitem[Shannon, 1948]{shannon1948mathematical}
Shannon, C.~E. (1948).
\newblock A mathematical theory of communication.
\newblock {\em Bell System Technical Journal}, 27:379--423.

\bibitem[Skyrms, 2010]{skyrms2010signals}
Skyrms, B. (2010).
\newblock {\em Signals: Evolution, Learning, and Information}.
\newblock Oxford University Press.

\bibitem[Talebirad et~al., 2026]{talebirad2026hierarchical}
Talebirad, Y., Parsaee, A., Szepesv\'{a}ri, C.~Y., Nadiri, A., and Za\"{i}ane,
  O.~R. (2026).
\newblock Toward a theory of hierarchical memory for language agents.
\newblock In {\em ICLR 2026 Workshop on Memory for LLM-Based Agentic Systems}.
\newblock arXiv preprint arXiv:2603.21564.

\bibitem[Tishby et~al., 1999]{tishby1999information}
Tishby, N., Pereira, F.~C., and Bialek, W. (1999).
\newblock The information bottleneck method.
\newblock In {\em 37th Annual Allerton Conference on Communication, Control and
  Computing}.

\bibitem[Wei et~al., 2022]{wei2022chain}
Wei, J., Wang, X., Schuurmans, D., Bosma, M., Ichter, B., Xia, F., Chi, E., Le,
  Q., and Zhou, D. (2022).
\newblock Chain-of-thought prompting elicits reasoning in large language
  models.
\newblock In {\em Advances in Neural Information Processing Systems}.

\bibitem[Zaslavsky et~al., 2018]{zaslavsky2018efficient}
Zaslavsky, N., Kemp, C., Regier, T., and Tishby, N. (2018).
\newblock Efficient compression in color naming and its evolution.
\newblock {\em Proceedings of the National Academy of Sciences},
  115:7937--7942.

\end{thebibliography}

\end{document}